\documentclass[10pt, conference]{IEEEtran}

\usepackage[utf8]{inputenc}
\usepackage[T1]{fontenc}
\usepackage{times}                  
\usepackage{cite}                   
\usepackage{microtype}              
\usepackage[inline]{enumitem}
\usepackage{amsmath}                
\usepackage{amssymb}                
\usepackage{amsthm}                 
\usepackage{mathtools}              
\usepackage{bm}                     
\usepackage{xfrac}                  

\usepackage[ruled, vlined]{algorithm2e}    

\usepackage{graphicx}               
\usepackage{subcaption}             
\usepackage{float}                  

\usepackage[dvipsnames]{xcolor}     

\usepackage{booktabs}               
\usepackage{multirow}               
\usepackage{multicol}               
\usepackage{array}                  

\usepackage{url}                    

\usepackage{enumitem}               

\usepackage[disable]{todonotes}
\usepackage[colorlinks=true, linkcolor=blue, citecolor=blue, urlcolor=blue]{hyperref}
\makeatletter
\let\ps@IEEEtitlepagestyle\ps@plain
\makeatother
\definecolor{independent}{RGB}{231,76,60}
\definecolor{centralized}{RGB}{46,204,113}
\definecolor{collaborative}{RGB}{52,152,219}
\definecolor{cbocu}{RGB}{155,89,182}
\definecolor{cbocl}{RGB}{230,126,34}

\DeclareRobustCommand{\methodline}[1]{%
    \raisebox{0.5ex}{\textcolor{#1}{\rule{1.2em}{1.2pt}}}%
}

\title{Privacy-Aware Collaborative and
Distributed Bayesian Optimization}
\author{
    \IEEEauthorblockN{
        Aditya Rane\textsuperscript{*}, Sathwik Yamana, Paritosh Ramanan, Srikanthan Ramesh, Akash Deep\textsuperscript{*}
    }
    \IEEEauthorblockA{
        School of Industrial Engineering and Management\\
        Oklahoma State University, Stillwater, OK, USA\\
        Email: aditya.rane10@okstate.edu; akash.deep@okstate.edu\\
    }
}
\begin{document}

\maketitle
\thispagestyle{plain}


\begin{abstract}
We propose a collaborative meta-learning framework for distributed Bayesian optimization matching centralized performance without raw-data exchange. We show gradient sharing leaks client observations, with leakage worsening as the search converges and queries concentrate near the optimum. We evaluate a differentially private defense and characterize its privacy-utility trade-off.
\end{abstract}
\begin{IEEEkeywords}
 Collaborative Optimization, Meta-Learning, Differential Privacy, Manufacturing Process Optimization
\end{IEEEkeywords}
\section{Introduction}
\label{sec:intro}
Modern scientific and manufacturing systems are increasingly distributed, operating across multiple geographic locations, research laboratories, and distinct organizational entities. In applications such as optimizing drug formulations, tuning additive manufacturing parameters, refining semiconductor processes, and accelerating new material discovery, independent facilities or clients aim to optimize related processes under different operating conditions \cite{chen2024exploring}\cite{raccuglia2016machine}. In practice, manufacturers rely on repeated Design of Experiments (DOE) studies and iterative build-inspect-adjust loops to establish stable process windows for every new configuration \cite{pugliese2026materials}. These cycles significantly lengthen development timelines, increase material usage, and constrain the overall throughput of advanced manufacturing. 

Underlying these inefficiencies is that process knowledge remains isolated across manufacturing facilities. Because engineers at one facility cannot access the optimization data or insights generated by another, they are forced to "start from scratch" for each new objective, unable to benefit from the collective experience, exists across the broader manufacturing ecosystem. Overcoming this limitation requires collaborative learning methods capable of pooling distributed information to extract shared structural knowledge, an inductive bias across related processes. However, due to proprietary concerns, trade secrets, or regulatory constraints, sharing the raw process data required to build these joint models is prohibited \cite{yang2019federated}. 

Bayesian optimization (BO) has proven highly sample-efficient for optimizing expensive, black-box manufacturing processes by constructing probabilistic surrogate models to guide sequential decisions \cite{frazier2018tutorial}. Standard BO frameworks, however, require raw data to be pooled in a single location, forcing facilities into either design isolation or mandatory data sharing. Recent advances in meta-learning, such as the PAC-Optimal Hyper-Posterior (PACOH) framework, enable extraction of a global inductive bias across related tasks but they assume a single, centralized entity with access to all historical task data, again necessitating the sharing of confidential information \cite{rothfuss2023scalable}. Extending these meta-learning models to a collaborative, decentralized setting by exchanging task-specific knowledge in the form of gradients, rather than raw data, introduces severe and underexplored privacy risks. Specifically, communicating these gradients to a central coordinator exposes the system to adversarial exploitation, intercepted gradients can be leveraged in Deep Leakage from Gradients (DLG) attacks to reconstruct the proprietary training data, the distributed architecture was designed to protect \cite{zhu2019deep}. 

Motivated by this challenge, we present in this paper how Privacy-Aware Collaborative and Distributed Bayesian Optimization (PACD-BO), enables independent facilities to jointly learn a shared optimization prior without exchanging raw data. Extending PACOH meta-learning, PACD-BO distributes the meta-prior update across clients through gradient exchange, approaching centralized performance while keeping observations local. In studying this collaborative setting, we further address BO specific privacy vulnerability and demonstrate differentially privacy as a potential defense mechanism.
\section{Related Work}
\label{sec:related}
\subsection{Knowledge Transfer in Sequential Design}
In advanced manufacturing, physical experiments are resource-intensive and strictly bounded by experimental budgets. While BO provides a sample-efficient alternative for expensive black-box optimization \cite{frazier2018tutorial}\cite{shahriari2015taking}, it inherently suffers from a cold-start problem, requiring many evaluations for every new task configuration. In decentralized settings, this cost is compounded by design isolation, because facilities cannot share raw data, each cold start its own processes from scratch, unable to exploit related experience. Meta-learning overcomes this bottleneck by transferring shared structure across related processes. State-of-the-art frameworks such as PACOH learn a meta-prior with PAC-Bayesian guarantees, and F-PACOH extends this regularization to the function space, treating meta-learned priors as stochastic processes for calibrated uncertainty quantification \cite{rothfuss2023scalable}\cite{rothfussfpacoh}. However, these methods are formulated for a centralized learner and do not address the privacy exposure introduced when task knowledge is shared across multiple clients.
\subsection{Collaborative and Distributed Optimization}
The centralization limitation above has motivated work on collaborative optimization, in which multiple clients leverage collective intelligence without centralizing their data. Federated learning achieves decentralized modeling by sharing local updates rather than raw data \cite{mcmahan2017communication}, a data-minimization principle that has recently motivated a range of collaborative BO methods. Some notable architectures include Collaborative BO via consensus (CBOC) \cite{yue2025collaborative}, which synchronizes agent proposals via a communication matrix, Collaborative Contextual BO (CCBO) \cite{chang2026collaborative}, which aggregates compressed posterior representations across clients, and Wasserstein-barycenter consensus \cite{zhan2025collaborative}, which fuses agents local GP surrogates into a central GP. While these methods demonstrate the utility of distributed collaboration, they treat the transmitted parameters such as proposals, posterior means, or covariances as privacy preserving, without a formal adversary model or empirical attack evaluation. Even \cite{zhan2025collaborative}, which acknowledges that the shared posterior statistics are sufficient  to reconstruct local data, relied on spatial discretizations as a defense. Consequently, reconstruction vulnerabilities such as gradient leakage remain largely unexamined in adversarial settings.
\subsection{Gradient Leakage in Sequential Optimization}
These reconstruction risks are well documented in other machine learning settings. DLG and its variants demonstrate that shared gradients, though assumed to be a secure form of data transmission, can be inverted to recover design data \cite{zhu2019deep}\cite{fernandez2025differential}. The standard defense, DP-SGD, clips per-example gradients and adds calibrated Gaussian noise, with the privacy budget tracked via Renyi accounting \cite{abadi2016deep}\cite{li2019differentially}. However, these attacks and defenses have been studied primarily in standard classification, their implications remain underexplored for BO, particularly within decentralized, coordinator–client architectures. In such settings, the sequential BO queries continuously adapt to sample the parameter space, and the resulting multiple communication rounds create a distinct, evolving privacy attack surface. Because this leakage differs fundamentally from that of standard deep learning, it demands privacy guarantees that prevent reconstruction without degrading optimization quality.   

Current literature lacks a secure architecture for distributed BO and neglects
the gradient leakage arising from sequential, acquisition-driven queries. We
address this with PACD-BO, which contributes
\begin{enumerate*}[label=(\roman*), itemjoin={{; }}, itemjoin*={{; and }}]
    \item an MPI-based framework that scales meta-learned BO across clients
    without raw-data exchange
    \item an empirical analysis adapting Deep Leakage from Gradients (DLG)
    attacks to the sequential setting, revealing a BO-specific clustering
    vulnerability
    \item a task-level differentially private mechanism with calibrated
    Gaussian noise, together with an analysis of its privacy--utility trade-off
\end{enumerate*}.
\subsection{Overview of the paper}
The paper is organized as follows, Section \ref{sec:framework} formalizes the decentralized collaborative framework, which is evaluated empirically in Section \ref{sec:performance}. Section \ref{sec:privacy} introduces the gradient leakage threat model, followed by our differentially private defense in \ref{sec:defense}. Finally, Section~\ref{sec:discussion} concludes the paper and outlines directions for future work.

\section{Collaborative Online PACD-BO Framework}
\label{sec:framework}
\begin{figure}[ht]
    \vspace{-3mm} 
    \centering
    \includegraphics[width=1\columnwidth]{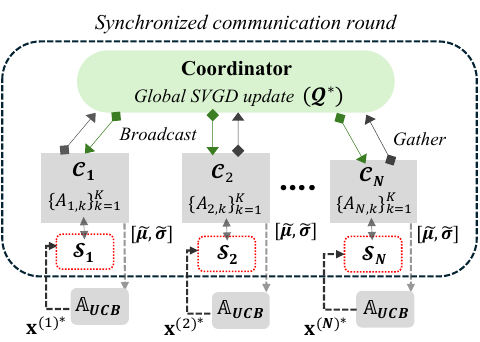}
    \vspace{-3mm} 
    \caption{Overview of PACD-BO collaborative framework. The coordinator broadcasts the global level $\mathcal{Q}^*$ and aggregates client gradients $\{A_{1,k}\}_{k=1}^K$ for the SVGD update. Each client $\mathcal{C}_i$ builds a local surrogate from $\mathcal{Q}^*$, selects an optimal query $\mathbf{x}^{(i)*}$ via UCB, and evaluates it to update its private data $\mathcal{S}_i$}
    \label{fig:framework}
    \vspace{-5mm} 
\end{figure}
\subsection{Problem Formulation}
We consider $N$ distributed clients $\mathcal{C} = \{1, \ldots, N\}$, representing independent facilities such as geographically separated manufacturing sites or research laboratories. Each client $i$ operates a private black-box objective function $f_i: \mathcal{X} \to \mathbb{R}$ over design space $\mathcal{X} \subseteq \mathbb{R}^d$, representing a process-quality response as a function of design or process parameters $\mathbf{x} \in \mathcal{X}$, and stores a private initial dataset $\mathcal{S}_i^{(0)} = \{(\mathbf{x}_j^{(i)}, y_j^{(i)})\}_{j=1}^{m_i}$ of past evaluations. We assume the observation model with additive Gaussian noise. 
\begin{equation}
    y_j^{(i)} = f_i(\mathbf{x}_j^{(i)}) + \epsilon_j^{(i)}, \quad \epsilon_j^{(i)} \sim \mathcal{N}(0, \sigma_n^2), \quad i = 1, \dots, N
    \label{eq:observation}
\end{equation}
where $\epsilon_j^{(i)}$ is independent and identically distributed noise. The objectives $\{f_i\}_{i=1}^N$ are related but non identical, reflecting shared physical structure under heterogeneous operating conditions. Furthermore, each $f_i$ corresponds to a task $\tau_i \sim \mathcal{T}$, so the tasks share latent structure while differing in their specifications. Additionally, client and task levels coincide, and we therefore index both by $i = 1, \dots, N$. Each client seeks to minimize its own $f_i$ using as few expensive evaluations as possible via BO. The goal for each client is to identify the optimal design parameters, benefiting from shared structure across the other clients.  
\begin{equation}
    \mathbf{x}_i^* = \arg\min_{\mathbf{x} \in \mathcal{X}} f_i(\mathbf{x}), \quad i = 1, \dots, N
    \label{eq:objective}
\end{equation}
Additionally, this collaboration is subject to a strict privacy constraint, raw data $\mathcal{S}_i$ never leaves client $i$, clients may exchange only gradient information through a central coordinator.

The PACD-BO framework structures collaborative optimization into two nested loops (see Figure~\ref{fig:framework}). At the coordinator level, the primary objective is to learn a shared, global meta-prior without centralizing raw client data, obtained through synchronized gradient exchange. Each client computes a local score gradient on its private data, and the coordinator aggregates these gradients rather than the data itself. Data is stored locally; gradients are shared globally. At the client level, this globally aggregated knowledge acts as an informative prior to guide local sequential design. This structural separation establishes federated collaboration. However, as we show in Section~\ref{sec:privacy}, these shared gradients also leak each client's query locations.

\subsection{PACOH Background}
Meta-learning extracts shared inductive bias from a set of related tasks to enable rapid learning from limited observations. PACOH formulates this within a PAC-Bayesian framework; given $N$ related tasks with datasets $\{\mathcal{S}_{i}\}_{i=1}^N$, it learns a hyper-posterior $\mathcal{Q}$ over priors $P$, regularized by a hyper-prior $\mathcal{P}$ to prevent meta-overfitting \cite{rothfuss2023scalable}. Minimizing the PAC-Bayesian bound on transfer error yields the closed-form optimal hyper-posterior:
\begin{equation}
 \mathcal{Q}^{*}(P) \propto \mathcal{P}(P) \cdot \exp\left( \frac{\lambda}{N\beta+\lambda}\sum_{i=1}^{N} \ln Z_{\beta}(\mathcal{S}_{i},P) \right)   
\label{eq:optimal}
\end{equation}
where $Z_\beta(\mathcal{S}_i, P)$ is the generalized marginal likelihood (GMLL) of task $i$ under prior $P$, and $\beta, \lambda$ are task and meta-level regularization parameters. Since $\mathcal{Q}^*$ is intractable, it is approximated by $K$ particles $\{\phi_k\}_{k=1}^K$ parameterizing a prior $P_{\phi_k}$ and updated via Stein Variational Gradient Descent (SVGD), $\phi_k \leftarrow \phi_k + \eta \psi(\phi_k)$, where:
\[
    \psi(\phi_k) = \frac{1}{K} \sum_{k'=1}^K \left[ k(\phi_{k'}, \phi_k) \nabla_{\phi_{k'}} \ln \mathcal{Q}^*(\phi_{k'}) + \nabla_{\phi_{k'}} k(\phi_{k'}, \phi_k) \right]
\]
Here, $k(\cdot, \cdot)$ is an RBF kernel and $\eta$ is the step size. Substituting Eq.~\eqref{eq:optimal}, the centralized score function driving this update decomposes into a hyper-prior term and a linear sum of per-task gradients:
\begin{equation}
    \nabla_{\phi_k} \ln \mathcal{Q}^*(\phi_k) = \nabla_{\phi_k} \ln \mathcal{P}(\phi_k) + \frac{\lambda}{N\beta + \lambda} \sum_{i=1}^N A_{i,k}
    \label{eq:score_decomposition}
\end{equation}
where $A_{i,k} = \nabla_{\phi_k} \ln Z_\beta(\mathcal{S}_i, P_{\phi_k})$ represents the localized score contribution depending strictly on the private dataset $\mathcal{S}_i$. We instantiate the base learner as a Gaussian process (GP) with a neural network parameterized kernel \cite{rothfuss2023scalable}, mapping inputs into a learned feature space:
\begin{equation}
  k_\phi(x, x') = \frac{1}{2}\exp\!\left(-\tfrac{1}{2}\,\|\varphi_\phi(x) - \varphi_\phi(x')\|_2^2\right)
  \label{eq:nn_kernel}
\end{equation}
The particles $\{\phi_k\}$ represent the underlying neural network weights, providing calibrated posteriors for BO acquisition.

\subsection{Collaborative Framework}
The additive structure of the score function in Eq.~\eqref{eq:score_decomposition} is the basis of our collaborative framework. Because the task dependent term is a sum over independent per-task contributions $A_{i,k}$, each depending only on its own dataset $\mathcal{S}_i$, the global SVGD update can be computed without centralizing any raw data. Each task is assigned to a client that evaluates its own term locally and shares only the resulting gradient.

We execute this distributed computation through an iterative, two step communication protocol. First, at client level synchronized communication round, every client $i$ receives the current particles $\{\phi_k\}_{k=1}^K $, representing joint structural knowledge across participating clients $\mathcal{Q}^{*}$, broadcast by coordinator, and compute its localized score contribution on its private data.
\begin{equation}
    A_{i,k} = \nabla_{\phi_k} \ln Z_\beta(\mathcal{S}_i, P_{\phi_k}), \quad k = 1, \dots, K.
\end{equation}
The client transmits only the $K$ gradient vectors $\{A_{i,k}\}_{k=1}^K$ to the coordinator, its raw dataset $\mathcal{S}_i$ never leaves its local environment.
Second, at the coordinator level, the central server aggregates these localized score contributions from all $N$ clients to reconstruct the centralized score, and performs the SVGD update using the aggregated score $G_k = \sum_{i=1}^N A_{i,k}$, instead of the centralized sum. Substituting $G_k$, the centralized update from Eq.~\eqref{eq:score_decomposition} is reformulated for the coordinator as follows:
\begin{equation}
    \nabla_{\phi_k} \ln \mathcal{Q}^*(\phi_k) = \nabla_{\phi_k} \ln \mathcal{P}(\phi_k) + \frac{\lambda}{N\beta + \lambda} G_k
    \label{eq:svgd_aggregated}
\end{equation}
The updated particles are then broadcast back to all clients for the next round. Because Eq.~\eqref{eq:score_decomposition} is a linear sum over tasks, $G_k$ is mathematically equivalent to the centralized score, so distributing the computation across clients incurs no approximation. Consequently, collaborative PACD-BO recovers the centralized meta-update up to floating-point round-off, which explains why it closely matches centralized performance across our benchmarks (see Table~\ref{tab:collab_result} in Section~\ref{sec:performance}). After receiving the updated $\mathcal{Q}^*$, the client utilizes the learned hyper-posterior to execute a local BO. Each client uses its $K$ particles as informative prior to form a mixture of surrogates over corresponding GP posteriors (refer to \cite{rasmussen2003gaussian} for an introduction to GPs). The predictive distribution at new design point $\mathbf{x}^*$ is computed as:
\begin{align}
    \tilde{\mu}(\mathbf{x}) &= \frac{1}{K} \sum_{k=1}^K \mu_{\phi_k}(\mathbf{x}) \label{eq:pred_mean} \\
    \tilde{\sigma}^2(\mathbf{x}) &= \frac{1}{K} \sum_{k=1}^K \sigma_{\phi_k}^2(\mathbf{x}) + \frac{1}{K} \sum_{k=1}^K \left( \mu_{\phi_k}(\mathbf{x}) - \tilde{\mu}(\mathbf{x}) \right)^2 \label{eq:pred_variance}
\end{align}
where $\mu_{\phi_k}$ and $\sigma_{\phi_k}^2$ are the posterior mean and variance under the $k$-th particle prior $\phi_k$. The first term averages within-particle variance, while the second captures between-particle variance, providing the calibrated epistemic uncertainty. The client selects its next query point by maximizing the UCB acquisition function, $\mathbb{A}_{\text{UCB}}(\mathbf{x}) = \kappa\tilde{\sigma}(\mathbf{x}) - \tilde{\mu}(\mathbf{x})$, where $\kappa > 0$ controls the trade-off between exploitation ($\tilde{\mu}$) and exploration ($\tilde{\sigma}$). 
\begin{equation}
    \mathbf{x}_{t+1}^{(i)*} = \arg \max_{\mathbf{x} \in \mathcal{X}} \mathbb{A}_{\text{UCB}}(\mathbf{x}; \tilde{\mu}, \tilde{\sigma}), \quad i = 1, \dots, N
    \label{eq:ucb_selection}
\end{equation}
The client then evaluates its private objective to obtain $y_{t+1}^{(i)}$ and appends the observation to its local dataset, $\mathcal{S}_i \leftarrow \mathcal{S}_i \cup \{(\mathbf{x}_{t+1}^{(i)*}, y_{t+1}^{(i)})\}$. The updated dataset guides the next synchronized round, and the two-level loop repeats. This process continues until a stopping criterion is met, such as convergence or exhaustion of the experimental budget $T$.

\subsection{Message Passing Interface Implementation}
\label{sec:mpi}

We implement the collaborative framework as a distributed system using the Message Passing Interface (MPI) \cite{mpi4py}. The network is mapped to a coordinator--client architecture within a single \texttt{MPI\_COMM\_WORLD} communicator: the coordinator runs on Rank $0$, and $N$ clients run on Ranks $1$ through $N$. Coordination is synchronous, and clients communicate only with the coordinator, never with one another. Each SVGD step constitutes one communication cycle consisting of three phases. \textbf{Broadcast:} The coordinator transmits the SVGD particles $\boldsymbol{\Phi}$, representing the shared $\mathcal{Q}^*$, to all clients via \texttt{MPI\_Bcast}. A broadcast is used rather than \texttt{MPI\_Scatter} because every client requires the complete particle set, not a fragment, to evaluate its local surrogate. \textbf{Local computation:} Each client instantiates the neural-network-parameterized GP model, evaluates the GMLL on its local dataset $\mathcal{S}_i$, and computes its local score gradient $\{A_{i,k}\}_{k=1}^{K}$. These computations proceed in parallel across clients. \textbf{Aggregation and update:} Clients return their gradients via \texttt{MPI\_Gather}. The coordinator sums them into the aggregated score $G_k$, adds the hyper-prior term $\nabla_{\phi_k}\ln\mathcal{P}(\phi_k)$, applies the SVGD update, and broadcasts the updated particles for the next step.

\section{Empirical Performance Analysis}
\label{sec:performance}
\begin{figure}[ht]
    \centering
    \begin{minipage}[t]{0.48\linewidth}
        \centering
        \includegraphics[width=\linewidth]{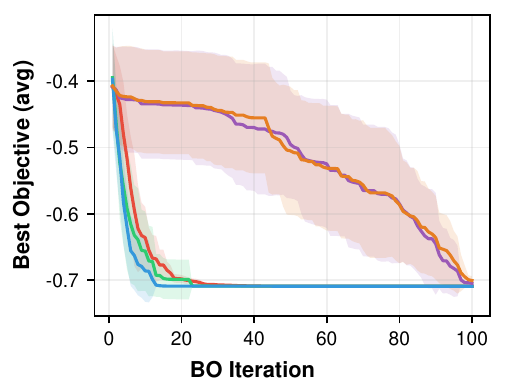}
        \vspace{1pt}
        {\small (a) Frequency Sinusoid (1D)}
    \end{minipage}
    \hfill
    \begin{minipage}[t]{0.48\linewidth}
        \centering
        \includegraphics[width=\linewidth]{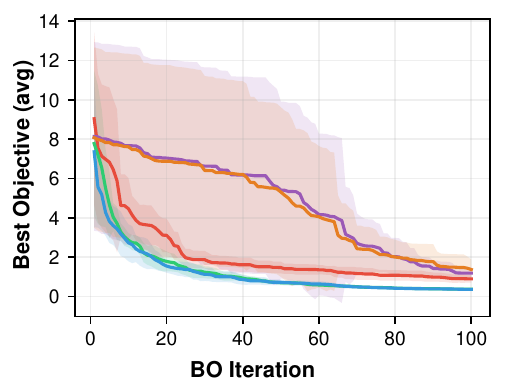}
        \vspace{1pt}
        {\small (b) Branin (2D)}
    \end{minipage}
    
    \vspace{1pt}
    
    \begin{minipage}[t]{0.48\linewidth}
        \centering
        \includegraphics[width=\linewidth]{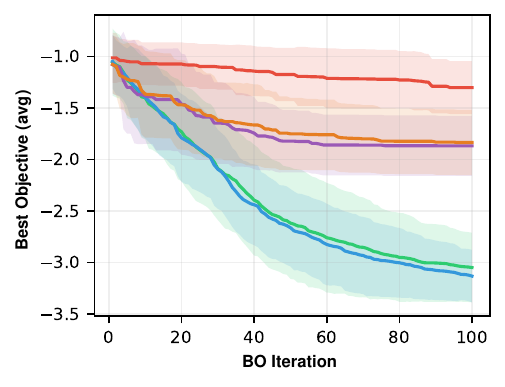}
        \vspace{1pt}
        {\small (c) Hartmann (6D)}
    \end{minipage}
    \hfill
    \begin{minipage}[t]{0.48\linewidth}
        \centering
        \includegraphics[width=\linewidth]{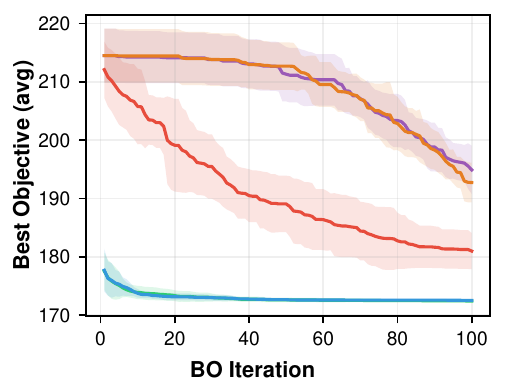}
        \vspace{1pt}
        {\small (d) Wing Weight (6D)}
    \end{minipage}
    \caption{Convergence analysis of the proposed PACD-BO against baseline architectures.
    \textbf{Methods:}\,
    \methodline{independent}\,Independent GPs,\,
    \methodline{centralized}\,PACOH-BO,\,
    \methodline{collaborative}\,PACD-BO,\,
    \methodline{cbocu}\,CBOC-U,\,
    \methodline{cbocl}\,CBOC-L}
    \label{fig:collab_res}
\end{figure}

\begin{table}[ht]
    \vspace{-2mm} 
    \centering
    \caption{Normalized regret (top) and AUC (bottom), mean $\pm$ std over 10 seeds (Lower is better)}
    \label{tab:collab_result}

    \setlength{\tabcolsep}{1.5pt} 
    \renewcommand{\arraystretch}{1.0} 

    \resizebox{\columnwidth}{!}{
    \begin{tabular}{lcccc}
    \toprule
    Method & Sin. & Bran. & Hart. & Wing \\
    \midrule
    \multicolumn{5}{l}{\textit{Normalized Regret}} \\
    Independent   & $\mathbf{.0000}_{\pm.0000}$ & $.0016_{\pm.0006}$ & $.5835_{\pm.0701}$ & $.0169_{\pm.0119}$ \\
    PACOH-BO      & $\mathbf{.0000}_{\pm.0000}$ & $\mathbf{.0000}_{\pm.0000}$ & $.0537_{\pm.1025}$ & $\mathbf{.0000}_{\pm.0000}$ \\
    \textbf{PACD-BO} & $\mathbf{.0000}_{\pm.0000}$ & $\mathbf{.0000}_{\pm.0000}$ & $\mathbf{.0300}_{\pm.0576}$ & $\mathbf{.0000}_{\pm.0000}$ \\
    CBOC-U        & $.0030_{\pm.0015}$ & $.0025_{\pm.0007}$ & $.4072_{\pm.1055}$ & $.0772_{\pm.0185}$ \\
    CBOC-L        & $.0058_{\pm.0044}$ & $.0033_{\pm.0018}$ & $.4168_{\pm.1174}$ & $.0693_{\pm.0164}$ \\
    \midrule
    \multicolumn{5}{l}{\textit{AUC}} \\
    Independent   & $.1194_{\pm.0350}$ & $.0185_{\pm.0069}$ & $.6620_{\pm.0534}$ & $.1451_{\pm.0288}$ \\
    PACOH-BO      & $.0899_{\pm.0422}$ & $.0153_{\pm.0053}$ & $\mathbf{.5758}_{\pm.1066}$ & $.0017_{\pm.0040}$ \\
    \textbf{PACD-BO} & $\mathbf{.0751}_{\pm.0284}$ & $\mathbf{.0133}_{\pm.0052}$ & $.5766_{\pm.1105}$ & $\mathbf{.0009}_{\pm.0025}$ \\
    CBOC-U        & $.1908_{\pm.0520}$ & $.0242_{\pm.0155}$ & $.5806_{\pm.1244}$ & $.1746_{\pm.0253}$ \\
    CBOC-L        & $.1925_{\pm.0505}$ & $.0239_{\pm.0148}$ & $.5921_{\pm.1060}$ & $.1750_{\pm.0256}$ \\
    \bottomrule
    \end{tabular}
    }
    \vspace{-4mm} 
\end{table}
\subsection{Experimental Setup}
We evaluate PACD-BO on four benchmarks with input dimensions $d \in \{1, 2, 6\}$: a frequency-varying sinusoid, Branin, Hartmann, and Wing Weight, all drawn from the Virtual Library of Simulation Experiments \cite{surjanovic2025virtual}. Each function is distributed across $N=5$ clients, with heterogeneity induced via task-defining parameters so that client objectives are related but distinct in shape, optima location, or operating conditions. 

Wing Weight represents our most application-relevant setting; it models a light aircraft wing as a function of ten variables. Each client optimizes six design variables (wing area, aspect ratio, quarter-chord sweep, taper ratio, thickness-to-chord ratio, and paint weight), while the remaining four are fixed at client-specific operating conditions (fuel weight, dynamic pressure, ultimate load factor, and flight design gross weight). This setup represents distinct manufacturing facilities running identical processes under different operational constraints. Every client is initialized with $5$ points for $1\text{D}$ and $2\text{D}$ functions and $10$ points for $6\text{D}$ functions, sampled via collaborative Latin hypercube sampling \cite{yue2025collaborative}. Each client runs $T=100$ BO iterations, with results averaged over $10$ independent replications.

We compare collaborative PACD-BO against five baseline configurations. \textit{Independent} runs standard BO at each client with no information sharing, establishing a design-isolation baseline. Conversely, the centralized reference (PACOH-BO) pools all client data into a unified meta-learner, serving as the ideal baseline we aim to match without sharing raw data. Finally, CBOC-U and CBOC-L are consensus-based collaborative BO variants \cite{yue2025collaborative} utilizing uniform and leader-weighted communication matrices, respectively. To evaluate performance under constrained budgets, we report normalized final regret alongside normalized AUC \cite{grosnit2021we} computed strictly over the first $10\%$ of iterations to measure rate of early convergence.
\begin{figure*}[t]
  \centering
  \includegraphics[width=\textwidth]{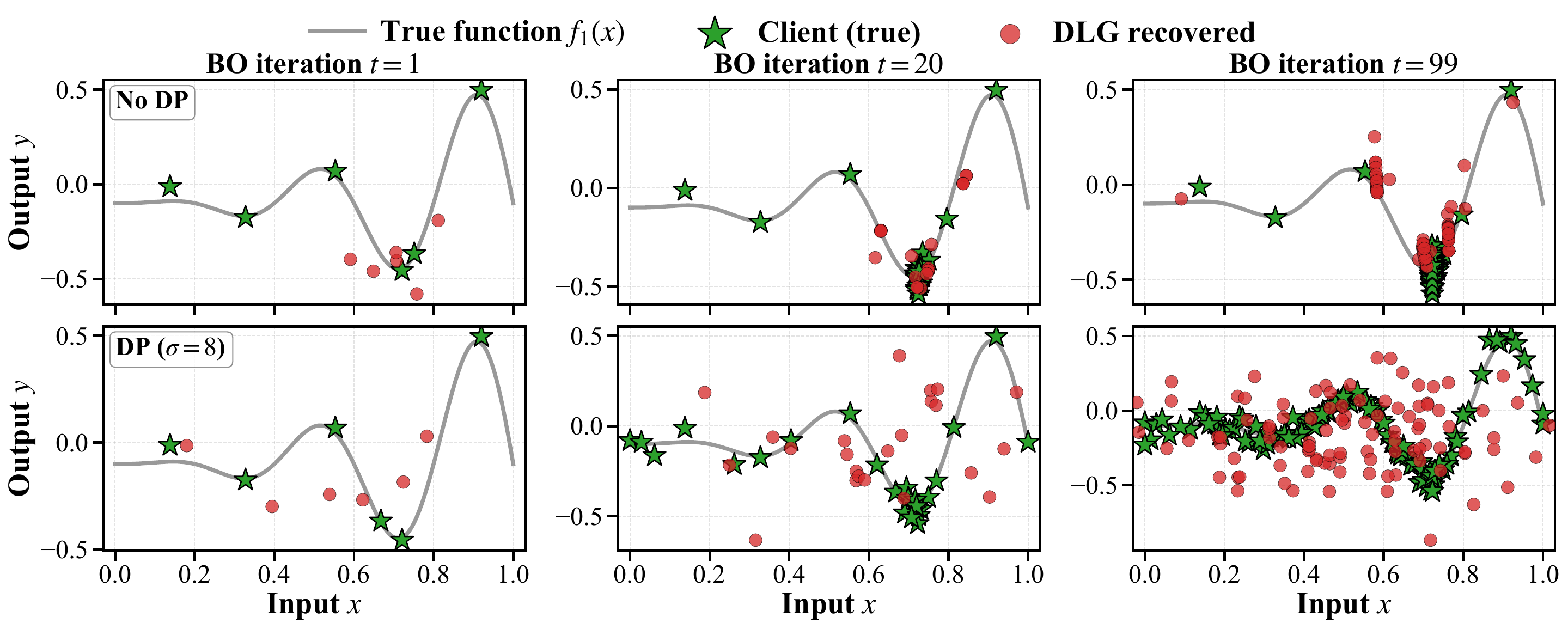}
    \caption{Spatial gradient leakage and defense. Unprotected gradients (Top) allow reconstruction of $\mathcal{C}_1$'s  queries by $t=99$, whereas differential privacy ($\sigma=8$, Bottom) prevents recovery of the true search trajectory.}
  \label{fig:spatial_grid}
\end{figure*}
\subsection{Results}
The performance across all evaluated benchmarks is shown in Table~\ref{tab:collab_result} and Figure~\ref{fig:collab_res}. PACD-BO empirically matches centralized meta-learning. Our framework performs equivalently to the centralized PACOH-BO baseline across all benchmarks, achieving $0$ normalized regret on sinusoid, Branin, and Wing Weight, and remaining within one standard deviation on Hartmann. Crucially, it also improves substantially over Independent BO in which each site optimizes alone. By communicating only gradient updates, PACD-BO maintains raw-data isolation at the client level without sacrificing optimization performance. Furthermore, consensus methods degrade under task heterogeneity. In terms of final regret, PACD-BO outperforms both CBOC variants on every benchmark, on Wing Weight, both CBOC variants underperform even Independent BO. Additionally, on Hartmann, CBOC's regret exceeds ours by a factor of $14$ ($.41$--$.42$ vs.\ $.030$). This indicates averaging proposals in a consensus framework distorts search when client optima diverge, whereas PACD-BO's joint meta-prior permits convergence to task-specific optima. On Hartmann, early-stage AUC does not discriminate among the collaborative methods all fall within one standard deviation so final regret is the informative metric.
\section{Privacy Analysis}
\label{sec:privacy}

\subsection{Threat Model and Attack Formulation}
We consider an honest-but-curious central coordinator that intercepts the localized score gradients $A_{i^*,k}^{(t)}$ from a target client $i^*$ to reconstruct its private dataset $\mathcal{S}_{i^*}^{(t)} = \{(x_j, y_j)\}_{j=1}^{m_t}$. The adversary adapts a DLG attack to the sequential BO setting by initializing dummy data $(\hat{X}, \hat{y}) \sim \mathcal{N}(0, I)$ \cite{zhu2019deep}. Let $\tilde{g}_k(\hat{X}, \hat{y})$ denote the dummy gradient evaluated on the broadcasted hyper-posterior particles. The adversary updates the dummy parameters using an L-BFGS optimizer to minimize the squared $\ell_2$ gradient matching loss:
\begin{equation}
\mathcal{L}_{\text{DLG}}^{(t)}(\hat{X}, \hat{y}) = \sum_{k=1}^K \Vert \tilde{g}_k(\hat{X}, \hat{y}) - A_{i^*,k}^{(t)} \Vert_2^2
\label{eq:dlg_loss}
\end{equation}
To avoid local minima and ensure robust recovery, the adversary executes a multi-start strategy across $R$ random initializations, selecting the optimized dataset that minimizes $\mathcal{L}_{\text{DLG}}^{(t)}$.
\subsection{Attack Results and Exploitation Leakage}
Unlike standard DLG, which inverts a single gradient, the PACD-BO exposes the target across sequential synchronized rounds. To evaluate this vulnerability, we simulated the attack on $\mathcal{C}_1$ across targeted BO iterations ($t = 1,5,20,40,60,80, 99$) utilizing 5 random restarts per round. This reveals a privacy phenomenon unique to BO, which we term ``Exploitation Leakage'' (Figure.~\ref{fig:spatial_grid}, top). During the initial exploration phase ($t=1$), the $\mathcal{C}_1$ queries are distributed broadly across the parameter space.  However, as optimization converges ($t=99$), queries tightly cluster at the true optimum. We identify that this spatial density severely worsens gradient leakage. The clustering of queried points allows the adversary’s optimizer to perfectly align the dummy points with the $\mathcal{C}_1$'s true queries, thereby exposing the most sensitive, highly optimized proprietary parameters.

\section{Differentially Private Defense}
\label{sec:defense}
\begin{figure}[ht]
  \centering
  \vspace{-2mm}                  
  \includegraphics[width=0.495\columnwidth]{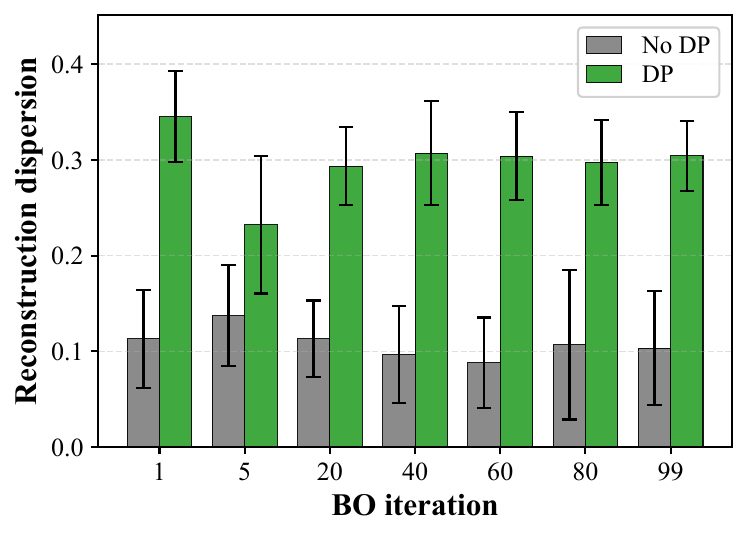}%
  \includegraphics[width=0.495\columnwidth]{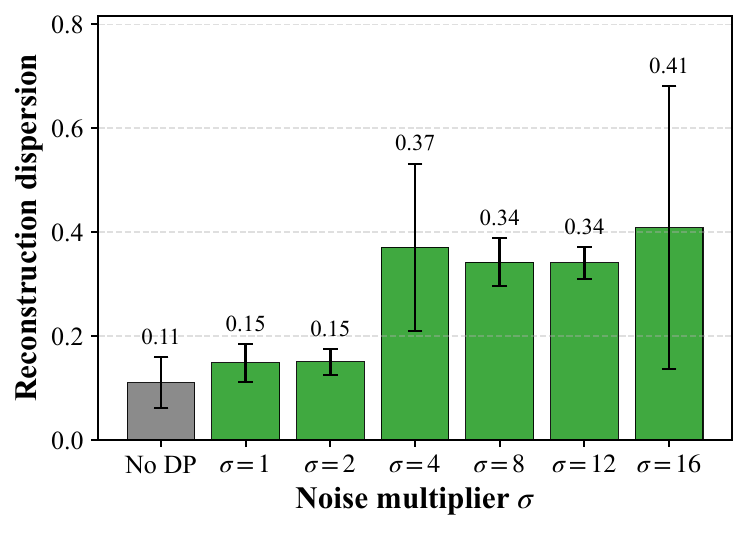}%
  \vspace{-3mm}                  
  \caption{Reconstruction dispersion analysis. Left: Dispersion across BO iterations. Right: Sensitivity across $\sigma$ multipliers.}
  \label{fig:dispersion}
  \vspace{-5mm}                  
\end{figure}

We defend against exploitation leakage with a task-level differentially private mechanism applied to each client's gradient before it leaves the client. The mechanism follows the standard Gaussian approach,
adapted to the per-particle gradients of the SVGD update~\cite{abadi2016deep}.
Each client first bounds the contribution of its update by clipping every
per-particle gradient to a fixed $\ell_2$ norm $C$:
\begin{equation}
    \bar{A}_{i,k}^{(t)} = A_{i,k}^{(t)} \cdot
    \min\!\left(1, \frac{C}{\|A_{i,k}^{(t)}\|_2}\right).
    \label{eq:clip}
\end{equation}
Clipping bounds the sensitivity of a single particle-gradient to $\Delta = 2C$, so that no individual update can dominate the aggregated score. Each clipped gradient is then perturbed with independent
Gaussian noise:
\begin{equation}
    \tilde{A}_{i,k}^{(t)} = \bar{A}_{i,k}^{(t)} + \mathbf{Z}_{i,k}, \qquad
    \mathbf{Z}_{i,k} \sim \mathcal{N}(0,\, \sigma^2 \Delta^2 \mathbf{I}),
    \label{eq:dp_noise}
\end{equation}
drawn independently for each particle $k$, where $\sigma$ is the noise
multiplier. The client transmits only the privatized gradients
$\{\tilde{A}_{i,k}^{(t)}\}_{k=1}^{K}$. Because the coordinator's aggregation and
SVGD update operate on these privatized quantities, they introduce no further leakage. Next, we select the $\sigma$ using Figure.~\ref{fig:dispersion} (right), which reports reconstruction dispersion (mean pairwise distance among DLG-recovered points), where higher values indicate a more scattered, less informative reconstruction across $\sigma$ multipliers. Dispersion saturates for
$\sigma \geq 4$, but $\sigma=4$ and $\sigma=16$ exhibit high variance across
seeds, indicating that their protection is unreliable. In contrast, $\sigma=8$
attains consistent dispersion with the smallest noise magnitude in the stable
regime, and we adopt it as our operating point. Figure.~\ref{fig:dispersion} (left) confirms that,
without DP, the recovered points remain tightly clustered (dispersion
$\approx 0.10$) at every BO iteration, whereas with DP maintains high
dispersion ($\approx 0.30$), dispersing the reconstruction across
the entire optimization trajectory.
\begin{figure}[ht]
  \centering
  \vspace{-2mm}                  
  \includegraphics[width=0.5\columnwidth]{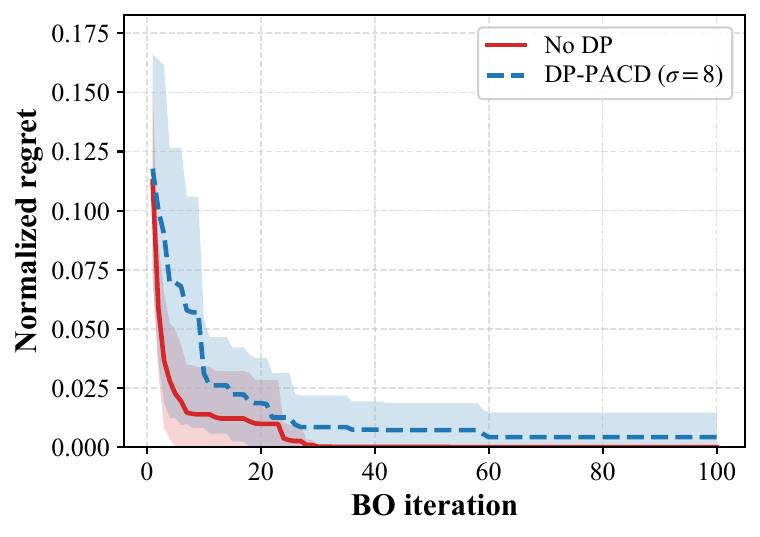}%
  \includegraphics[width=0.5\columnwidth]{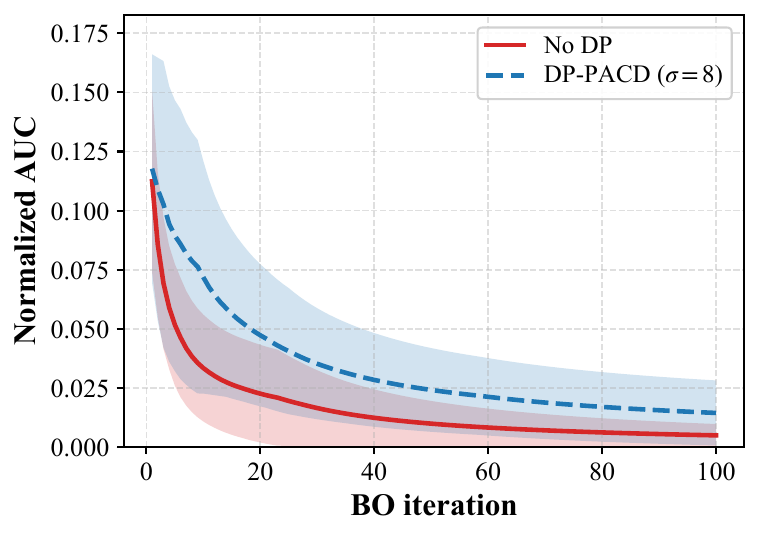}%
  \vspace{-3mm}                  
  \caption{Privacy-utility trade-off. Optimization performance with DP ($\sigma=8$). DP noise slows down the early stage convergence.}
  \label{fig:dp_utility_cost}
  \vspace{-4mm}                  
\end{figure}

Finally, this protection comes at a cost to optimization, shown in Figure~\ref{fig:dp_utility_cost}, which reports normalized regret (left) and AUC (right). The injected noise slows early convergence, so DP-PACD-BO incurs higher regret around the first $30$ iterations. As
optimization proceeds the gap narrows, and by $t=100$ DP-PACD-BO reaches
near parity with the non-private model, with only a small residual gap that
lies within the run-to-run variability of both methods. The AUC gap persists by construction, since AUC is computed over the early window where the noise has its largest effect. Task-level DP 
obscures the client's search trajectory at a cost to optimization quality.

\section{Discussion and Conclusion}
\label{sec:discussion}
We presented PACD-BO, a collaborative framework for PAC-Bayesian meta-learning without centralizing raw data. By exploiting the additive PACOH score structure, PACD-BO reproduces the centralized update while outperforming independent and consensus-based baselines. To the best of our knowledge, this is the first study to demonstrate gradient-inversion attacks in a BO setting. 

Our findings show that an honest-but-curious coordinator can reconstruct client queries, with leakage worsening as BO converges near the optimum. Task-level DP disperses these reconstructions but slows early convergence. Future work will evaluate larger client populations, high-dimensional multi-objective settings, real manufacturing problems, more advanced gradient-inversion attacks, and BO-specific privacy mechanisms with adaptive clipping and noise schedules.

\bibliographystyle{unsrt}
\bibliography{references}

\end{document}